\pdfoutput=1

\documentclass[11pt]{article}

\usepackage[preprint]{acl}
\usepackage{times}
\usepackage{latexsym}

\usepackage[T1]{fontenc}

\usepackage[utf8]{inputenc}

\usepackage{microtype}

\usepackage{inconsolata}

\usepackage{graphicx}

\usepackage{booktabs} 
\usepackage{xcolor}   
\usepackage{colortbl} 
\usepackage{multirow} 
\usepackage{xspace}
\usepackage{listings}
\usepackage{booktabs}
\usepackage{amssymb}
\usepackage{amsmath} 

\usepackage{algorithm}
\usepackage{algorithmic}

\usepackage[capitalize]{cleveref}
\crefname{section}{Sec.}{Secs.}
\Crefname{section}{Section}{Sections}
\Crefname{table}{Table}{Tables}
\crefname{table}{Tab.}{Tabs.}
\Crefname{algorithm}{Algorithm}{Algorithms.}
\crefname{algorithm}{Alg.}{Algs.}

\usepackage{amsmath,amsfonts,bm}

\def\eqref#1{equation~\ref{#1}}

\def\1{\bm{1}}

\DeclareMathAlphabet{\mathsfit}{\encodingdefault}{\sfdefault}{m}{sl}
\SetMathAlphabet{\mathsfit}{bold}{\encodingdefault}{\sfdefault}{bx}{n}

\newcommand{\E}{\mathbb{E}}

\newcommand{\KL}{D_{\mathrm{KL}}}

\usepackage{hyperref}

\usepackage{enumitem}
\setenumerate{itemsep=0pt,partopsep=0pt,parsep=0pt,topsep=2pt}

\title{Logits-Based Finetuning}

\author{
  \textbf{Jingyao Li\textsuperscript{1}},
  \textbf{Senqiao Yang\textsuperscript{1}},
  \textbf{Sitong Wu\textsuperscript{1}},
  \textbf{Han Shi\textsuperscript{2}},
  \textbf{Chuanyang Zheng\textsuperscript{1}},
  \textbf{Hong Xu\textsuperscript{1}},
  \textbf{Jiaya Jia\textsuperscript{3}}
\\
\\
  \textsuperscript{1}The Chinese University of Hong Kong \\
  \textsuperscript{2}Huawei Noah’s Ark Lab \\
  \textsuperscript{3}Hong Kong University of Science and Technology 
}

\def\ourmodel{ScienceLLaMA\xspace}

\def\our{\text{L}}

\def\D{\mathcal{D}}

\def\KL{\mathcal{D}_{KL}}

\newcommand{\stu}[0]{p_{\text{S}}}
\newcommand{\tea}[0]{p_{\text{T}}}
\newcommand{\pgt}[0]{p_{\text{GT}}}
\newcommand{\para}[1]{\vspace{2mm} \noindent \textbf{#1}}

\begin{document}
\maketitle
\begin{abstract}
In recent years, developing compact and efficient large language models (LLMs) has emerged as a thriving area of research. Traditional Supervised Fine-Tuning (SFT), which relies on singular ground truth labels, often fails to capture  token-level dependencies and linguistic diversity. To address these limitations, we propose a logits-based fine-tuning framework that integrates the strengths of supervised learning and knowledge distillation. Our approach constructs enriched training targets by combining teacher logits with ground truth labels, preserving both correctness and linguistic diversity. This ensures more reliable and effective training. We constructed a large-scale 1.2M logits dataset and trained a series of science-focused models. Experimental results demonstrate that our method achieves significant improvements, with accuracy gains of 18\% on Mawps and 22.7\% on TabMWP. Across nine widely used mathematical benchmarks, our method consistently outperforms prior SFT models, achieving an average improvement of 7.28\%. 
Codes are available at \href{https://github.com/dvlab-research/Logits-Based-Finetuning}{https://github.com/dvlab-research/Logits-Based-Finetuning}.
\end{abstract}

\section{Introduction}
Large language models (LLMs) have demonstrated remarkable capabilities across a wide range of NLP tasks~\cite{gpt3,lamda,palm,gpt4,palm2}, yet their immense computational demands pose significant challenges for deployment in resource-constrained environments. 

To address this, researchers have focused on developing compact and efficient LLMs, with Supervised Fine-Tuning (SFT) as a widely adopted approach. However, SFT suffers from inherent limitations, particularly its inability to capture inter-token relationships and linguistic diversity. For instance, as illustrated in \cref{fig:framework}, multiple valid expressions of the same idea, such as "There are 12 inches in 1 foot" and "There are 12 inches in each foot," highlight the nuanced token-level dependencies that SFT often overlooks. This limitation stems from SFT’s reliance on singular ground truth labels or teacher outputs, which fail to account for the richness of alternative phrasings. Consequently, the benefits of SFT are constrained by its inability to fully exploit the intrinsic relationships between tokens.

\begin{figure}
    \centering
    \includegraphics[width=1\linewidth]{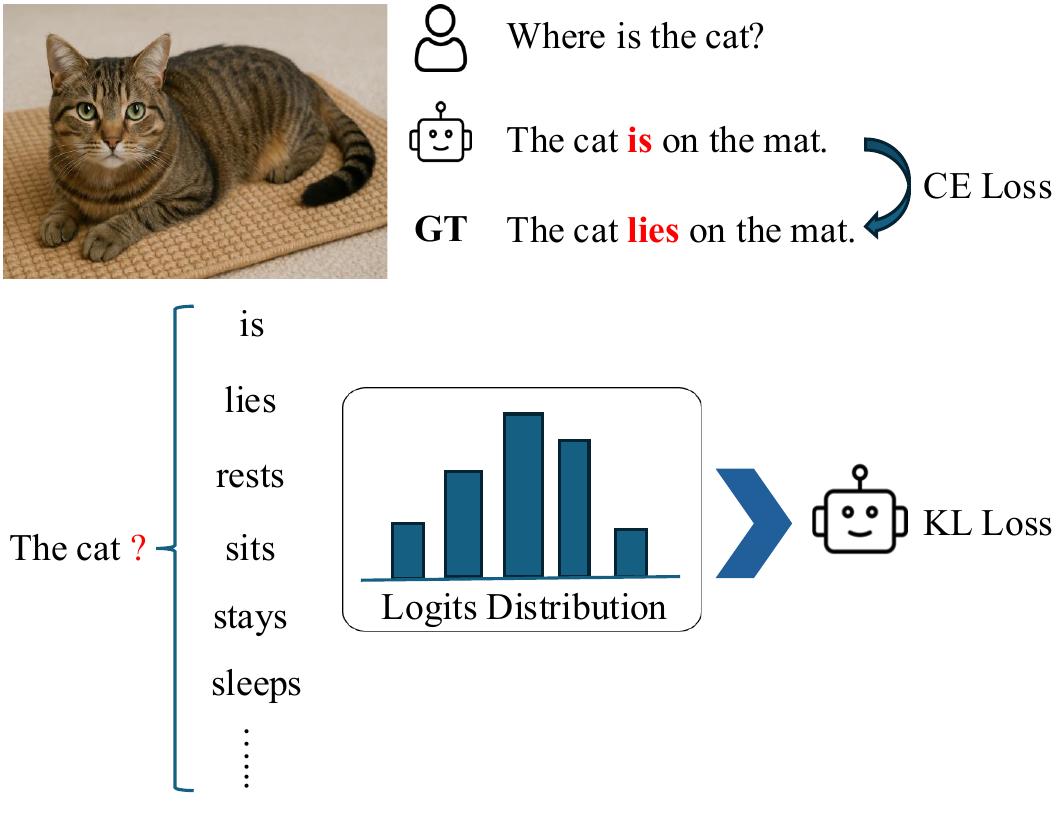}
    \vspace{-4mm}
    \caption{Conceptual overview of our logits-based distillation framework. (Up) Traditional supervised fine-tuning relies on singular ground truth labels, failing to capture valid linguistic variations (e.g., "The cat is on the mat" vs. "The cat lies on the mat"). (Down) Our approach combines teacher model logits with ground truth verification to create enriched training targets that preserve both correctness and expression diversity. }
    \label{fig:enter-label}
\end{figure}

Distillation methods have proven successful in creating lightweight and efficient models.  For example, models like BERT~\citep{policy_kd,distilbert,kdsurvey} have demonstrated that distillation-based approaches can achieve superior performance compared to direct training methods, offering both efficiency and effectiveness. However, applying distillation to LLMs presents unique challenges. First, the uncontrollability of teacher outputs poses a significant hurdle. Even well-trained large language models, such as LLaMA3.1-70B-instruct, can generate hallucinated or erroneous predictions, as shown in \cref{tab:wrong-teacher}. Relying solely on such outputs as supervision signals is unreliable and often necessitates manual intervention to ensure high-quality annotations. Second, the computational cost of large-scale distillation is prohibitive, as LLMs require substantial GPU memory, making direct online teacher-student distillation impractical for many applications.

To address these challenges, we propose a novel logits-based fine-tuning framework that integrates the strengths of supervised learning and knowledge distillation. Our approach constructs enriched training targets by combining teacher logits with ground truth labels, preserving both correctness and linguistic diversity. Unlike traditional distillation methods, which transfer teacher predictions directly, our method creates a balanced target distribution that enhances the student model’s ability to learn from both the teacher’s knowledge and task-specific supervision. This ensures more reliable and informed training while mitigating the risks associated with erroneous teacher outputs. 

In this work, we constructed a large-scale 1.2M logits dataset and trained a series of science-focused models using our method. Experimental results show that our approach surpasses the previous state-of-the-art methods on Mawps and TabMWP by 18\% and 22.7\% in accuracy, respectively. Across nine widely used mathematical benchmarks, our method consistently outperforms prior SFT models, with an average improvement of 7.28\%, highlighting the method’s robustness and generalizability.

In summary, the main contributions of our work are as follows:
\begin{enumerate}
    \item We propose a simple yet effective logits-based instruction tuning method that enhances model performance by integrating teacher knowledge with ground truth labels.
    \item We release a 1.2M science logits dataset, enabling future research and development of logits-based training methods.
    \item We train and evaluate a series of science-focused models using our method. Our models achieve significant improvements over state-of-the-art supervised fine-tuning approaches, with an average accuracy gain of 7.28\% across nine benchmarks.
\end{enumerate}

\begin{table*}[htpb]
    \centering
    \fontsize{9}{11.5}\selectfont 
    \begin{tabular}{p{0.45\textwidth} p{0.45\textwidth}}
    \multicolumn{2}{p{0.9\textwidth}}{\colorbox{orange!15}{\textbf{Instruction}}}\\
    \multicolumn{2}{p{0.9\textwidth}}{Roger's age is 5 more than twice Jill's age. In 15 years, their age difference will be 30 years less than Finley's age. If Jill is 20 years old now, how old is Finley?} \\
    \colorbox{teal!15}{\textbf{Teacher Annotations}} & \colorbox{teal!15}{\textbf{Human Annotations}} \\
Let's break down the problem step by step.

First, we know that Jill is 20 years old now. We also know that Roger's age is 5 more than twice Jill's age. So, Roger's age is $2 \times 20 + 5 = 45$.

In 15 years, Jill will be 20 + 15 = 35 years old, and Roger will be 45 + 15 = 60 years old. The problem states that their age difference will be 30 years less than Finley's age. So, we can set up the equation:

60 - 35 = Finley's age - 30

Simplifying the equation, we get:

25 = Finley's age - 30

Adding 30 to both sides, we get:

Finley's age = 55

Therefore, Finley is {\color{red}\textbf{55}} years old now. & 

Twice Jill's age is $2\times 20=40$.

If Roger's age is 5 more than twice Jill's age, he is 5+40=45 years old.

In 15 years, Jill will be 20+15=35 years old.

Roger's age will be 45+15=60 years old in 15 years.

The age difference between Roger and Jill in fifteen years will be 60-35=25 years.

In 15 years, the age difference between Roger and Jill will be 30 years less than Finley's age, meaning in fifteen years Finley will be 30+25=55 years old.

Currently, Finley is {\color{teal}\textbf{55-15=40}} years old.
\\
    \end{tabular}
    \caption{An example from GSM8K~\cite{gsm8k} generated by teacher model, LLaMA3.1-70B-instruct. The teacher model incorrectly interprets Finley's age 15 years from now as his current age, leading to an erroneous conclusion. }
    \label{tab:wrong-teacher}
\end{table*}

\section{Preliminaries}
\label{sec:preliminaries}
In this section, we establishes the theoretical foundation for our logits-based fine-tuning approach. We first formalize auto-regressive sequence modeling and then analyze existing knowledge distillation paradigms, highlighting their limitations that motivate our method.

\subsection{Auto-regressive Sequence Models}
We first define key components of sequence modeling. For any sequence pair, $x$ represents the input and $y$ the output. The vocabulary $\mathbb{V}$ contains $M$ distinct tokens. We use $y_{<n+1} = (y_1, y_2, \dots, y_{n})$ to represent the first $n$ tokens.
An auto-regressive model generates a probability distribution $p(. | y_{<n}, x) \in [0,1]^{M}$ over the vocabulary $\mathbb{V}$ , considering both input $x$ and previous tokens $y_{<n}$. When sampling, $y \sim p(\cdot|x)$ produces a complete output sequence. For brevity, we write $p(y_n|x)$ instead of $p(y_n | y_{<n}, x)$.
The generation process predicts tokens sequentially. Each token probability $p(y_n|x)$ is computed using a temperature-controlled softmax:
\begin{equation}
p(y_n|x) = \frac{e^{z_{n}/\tau}}{\sum_{i=1}^{M} e^{z_i /\tau}},   
\end{equation}
where $z_n$ represents the logit for token $y_n$, and $\tau$ controls output randomness. Higher $\tau$ increases diversity, while lower values produce more focused predictions. During student training, $\tau=1$, while evaluation uses greedy sampling ($\tau \rightarrow 0$).

\subsection{Distillation for Auto-regressive Models}
\paragraph{KL-Based Divergences.}
The Kullback-Leibler (KL) divergence is a fundamental measure that quantifies the difference between two probability distributions. For two discrete probability distributions $P(\cdot|x)$ and $Q(\cdot|x)$ defined over the probability space $X$, the KL divergence is defined as~\citep{kd}:
\begin{equation}
\KL(P|Q) = \sum_{x \in \mathcal{X}} P(\cdot|x) \log \frac{P(\cdot|x)}{Q(\cdot|x)}.
\end{equation}
This measure is always non-negative and equals zero if and only if the two distributions are identical.

\begin{figure*}[t]
  \includegraphics[width=\linewidth, trim=0 210 150 20, clip]{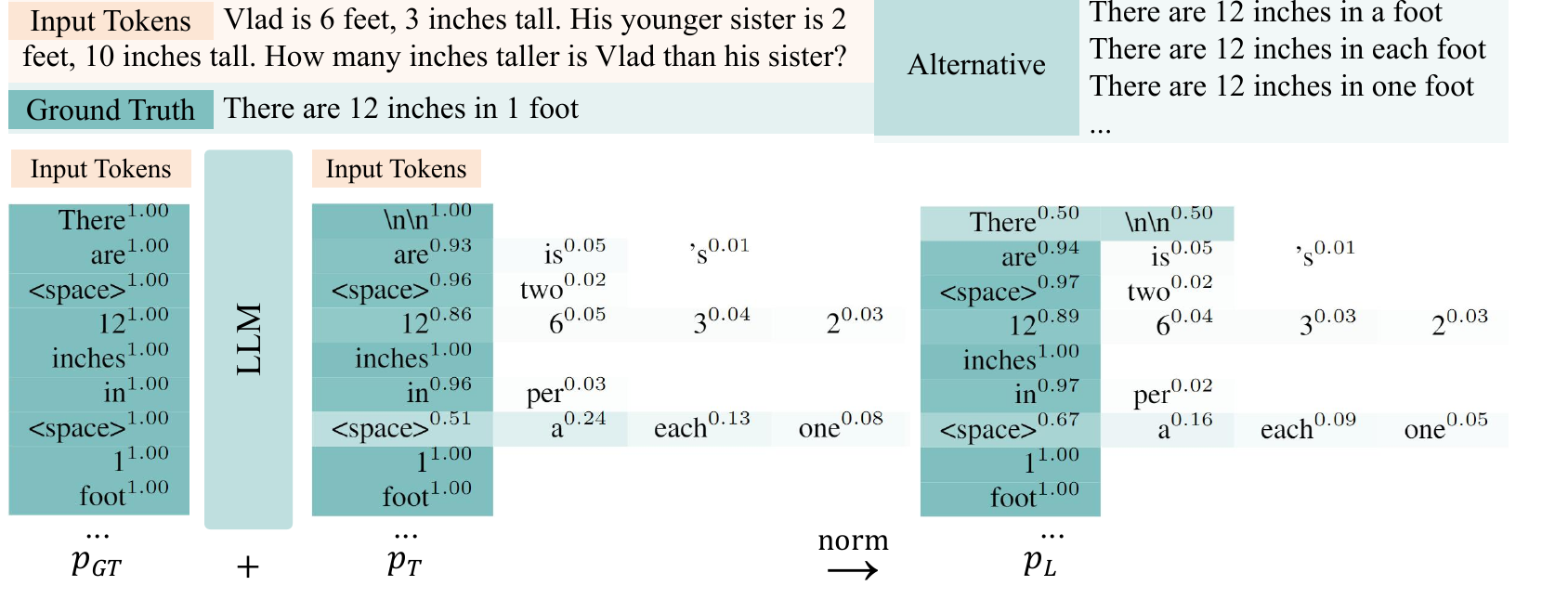}
  \caption{Illustration of token probability distribution generation. The input tokens concated with ground truth are processed by the teacher model, which predicts the next token probabilities \( p_T \). Then the ground truth one-hot vector \( P_{GT} \) is combined with the teacher's top-K probabilities \( p_T \) to generate the proposed distribution \( p_{\text{our}} \) using \cref{eq:pour}.  }
  \label{fig:framework}
\end{figure*}


\paragraph{Supervised FT.} Given a fixed dataset of target sequences, one simple strategy involves maximizing the student's negative log-likelihood on these sequences:
\begin{equation}\label{eq:sft}
L_{SFT}(\theta) = \E_{(x, y) \sim (X, Y)} \big[-\log\stu^\theta(y|x)\big].
\end{equation}

\paragraph{Sequence-Level KD} ~\citep{seqkd} extends this concept by training on teacher-generated outputs $Y_T$. This approach optimizes:
\begin{equation}\label{eq:seqkd}
L_{SeqKD}(\theta) = \E_{(x, y) \sim (X, Y_T)} \big[-\log\stu^\theta(y|x)\big].
\end{equation}

\paragraph{Supervised KD~}\citep{kd} represents a widely used distillation method where students learn to match their teacher's token-level probability distributions. The training objective minimizes the KL divergence between teacher and student distributions:
\begin{equation} \label{eq:sd}
L_{SD}(\theta) := \E_{(x, y) \sim (X, Y_T)} \Big[ \KL\big(\tea | \stu^\theta\big)(y|x)\Big],
\end{equation}

\section{Logits-based Finetuning}
In this section, we first introduce the motivation behind our logits-based fine-tuning approach in \cref{sec:motivation}. Then, in \cref{sec:constraints}, we present the proposed distribution, which integrates teacher model logits with ground truth outputs. In \cref{sec:lft_dataset}, we describe the construction of our logits dataset. Finally, in \cref{sec:lft_training}, we detail our fine-tuning method.

\subsection{Motivation}
\label{sec:motivation}
To justify the proposal of the Logits-Based Fine-Tuning method for improving small LLMs, we first analyze the limitations of traditional widely used method Supervised Fine-Tuning (SFT), and the current distillation method Sequence-Level Knowledge Distillation (SeqKD,~\citet{seqkd}), and Supervised Distillation (SD,~\citet{kd}):

\paragraph{Lack of Inter-Token Relationships.} For traditional SFT, the major issue is the lack of inter-token relationships. Specifically, there may be multiple expressions for the same idea, such as \textit{There are 12 inches in 1 foot} and \textit{There are 12 inches in each foot} illustrated in~\cref{fig:framework}. These alternative labels reflect the model's understanding of the intrinsic relationships between tokens, which may not be captured through singular annotations.

\paragraph{Uncontrollability of Teacher Outputs.} Besides, for the distillation method, the outputs from LLMs are often uncontrollable; even well-trained models can produce erroneous or hallucinatory results. For instance, as shown in~\cref{tab:wrong-teacher}, the well-trained LLaMA3.1-70B-instruct model erroneously interprets Finley's age 15 years from now as his current age, resulting in incorrect conclusions. Therefore, relying solely on the outputs of LLMs as supervision for models is unreliable and necessitates human intervention to generate validated results.

\subsection{Target Distribution Analysis}
\label{sec:constraints}
To address these limitations, we aim to propose a approach that enables the student model to learn from both reliably annotated labels and the intrinsic knowledge embedded in the teacher model.

\paragraph{Problem Setup.} Consider two sequence models with auto-regressive architectures: $\stu$ (student) and $\tea$ (teacher), with different model capacities. The student model has trainable parameters $\theta$, and $\stu^\theta$ maintains differentiability with respect to $\theta$. The setup includes an input dataset $X$.  We define the token-level distribution discrepancy between $\tea$ and $\stu$ as: 
\begin{equation}
\begin{aligned}
   & \D\big(\tea \| \stu^\theta\big)(y|x) \\
  & := 
  \frac{1}{L_y} \sum_{n=1}^{L_y} \D\big(\tea(\cdot|y_{<n},x) \| \stu^\theta(\cdot|y_{<n}, x)\big),\label{eq:div}
\end{aligned}
\end{equation}
where $x$ and $y$ denote the input and output sequences and $\D$ represents divergence measure.

\paragraph{Definition.}Let $M$ represent the vocabulary size and $y_i$ denote the $i$-th ground truth index, where $0 < y_i < M$. The target distribution is denoted as $q$. Specifically, $q_{j}(y_i)$ represents the value at the $j$-th position in the vocabulary for the $i$-th token's logits in the target logits $q$. Storing a vocabulary of millions of tokens incurs significant storage overhead. Therefore, we retain only the sparse teacher logits of the top $K$ instead of the complete set. For simplicity, all subsequent references to $\tea$ logits refer to the Top-K sparsified results. We define $\text{Top}_K\tea(y_i) = \text{Top}_{K,1 \le j \le M} \tea(y_i)$. 

\paragraph{Proposed Distribution.}
We propose our probability distribution $p_{\our}$ as follows:
\begin{equation} \label{eq:pour}
p_{\our}(y_i)=\frac{\tea(y_i)+\pgt(y_i)}{\|\tea(y_i)+\pgt(y_i)\|_1},
\end{equation}
where $\|\cdot\|_1$ denotes the L1 norm.  $\pgt(y_i)$ is the one-hot encoded ground truth label. Specifically, $\pgt(y_i)=\{p_{\text{GT}j}(y_i)\}_{j=1}^M \in [0,1]^{M}$, where
\begin{equation} \label{eq:onehot}
\pgt(y_i)^j=
\begin{cases}
1, & {\rm if } j = y_i, \\
0, & {\rm otherwise}.
\end{cases}
\end{equation}
We define this distribution because it satisfies the following constraints.

\paragraph{Constraint 1.} To ensure that the greedy search on the new distribution $q$ still yields the ground truth $y_i$, we require that the value $q(y_i)$ be the largest at the ground truth index. Mathematically, this is expressed as:
\begin{equation}
q_{y_i}(y_i) \ge q_{j}(y_i), \quad \forall 1\le j\le M, j \ne y_i
\end{equation}
This constraint guarantees that the argmax of $q(y_i)$ remains $y_i$, preserving the ground truth prediction.

\paragraph{Constraint 2.}
We aim to maintain the relative proportions of the top $K$ candidates from the original distribution $\tea(y_i)$ in the new distribution $q$. The constraint is formulated as:
\begin{equation}
\frac{q_j(y_i)}{q_k(y_i)} = \frac{\tea(y_i)_j}{\tea(y_i)_k}, \quad \forall j, k \in \text{Top}_K(y_i), j, k \ne y_i
\end{equation}
This ensures that the proportional relationship between the probabilities determined by the original distribution is preserved in the new distribution. 

\paragraph{Constraint 3.} 
For indices outside the ground truth and the top $K$ candidates, we require their values in $q$ to be not larger than those within the set $S = \{y_i\} \cup \text{Top}_K\tea(y_i)$. This is expressed as:
\begin{equation}
q_{j}(y_i) \le q_{k}(y_i), \quad \forall j \notin S, \forall k \in S
\end{equation}
This constraint helps in focusing the probability mass on the ground truth and the top candidates, reducing the influence of less relevant tokens.

\paragraph{Constraint 4.} 
Finally, the new distribution $q(y_i)$ must be a valid probability distribution. This implies that each element must be within the range [0, 1], and the sum of all elements must equal 1. Mathematically:
\begin{equation}
q(y_i) \in [0, 1]^M, \\
\sum_{j=1}^M q_{j}(y_i) = 1
\end{equation}
These constraints ensure that $q(y_i)$ is a well-formed probability distribution, suitable for logits-based fine-tuning. It can be easily demonstrated that $p_{\our}$ satisfies the four constraints outlined above. Details are in \cref{sec:demonstration}.

\begin{algorithm}[t]
\caption{Logits Dataset Generation Procedure}
\label{alg:lft_dataset}
\begin{algorithmic}[1]
\REQUIRE Teacher model $\tea$, Dataset $(X,Y) = {(x_i, y_i,)}_{i=1}^N$
\ENSURE Logits-based Dataset $(X,Y,P_{\our}) = {(x_i, y_i, p_{\our i})}_{i=1}^N$
\FOR{each $(x, y) \in (X,Y)$}
\STATE Compute Top-$K$ teacher logits $\tea \gets T(x)$
\STATE Create one-hot ground truth $\pgt$ using \cref{eq:onehot}.
\STATE Compute $p_{\our}$ using \cref{eq:pour}.
\ENDFOR
\RETURN Logits-based Dataset $(X,Y,P_{\our}) = {(x_i, y_i, p_{\our i})}_{i=1}^N$
\end{algorithmic}
\end{algorithm}

\begin{algorithm}[t]
\caption{Logits-based Finetuning Procedure}
\label{alg:lft_training}
\begin{algorithmic}[1]
\REQUIRE Student model $\stu^\theta$, Logits-based Dataset $(X,Y,P_{\our})$, Divergence $\D$, learning rate $\eta$
\ENSURE Trained student model $\stu^\theta$
\FOR{batch $B\in (X,Y,P_{\our})$}
\STATE Update student parameters $\theta$ by minimizing $L_{\our}$ (\cref{eq:lour}): $$ \theta \gets \theta - \eta \frac{1}{B} \sum_{(x, y, p_{\our}) \in B} \nabla_\theta \D(p_{\our} \| \stu^\theta) (y|x)$$
\ENDFOR
\RETURN Trained student model $\stu^\theta$
\end{algorithmic}
\end{algorithm}

\subsection{Logits Dataset Generation}
\label{sec:lft_dataset}
The logits dataset generation procedure, as detailed in \cref{alg:lft_dataset}, takes a standard dataset of input-target pairs and enriches it with target distributions derived from a pre-trained teacher model. 

For each input-target pair $(x, y)$, the teacher model $\tea$ is first used to compute the full logits vector for input $x$, which is then sparsified by retaining only the top $K$ logits, denoted as $\tea(x)$. 
This sparsification is crucial for reducing storage requirements and focusing on the teacher's most confident predictions. 
Concurrently, a one-hot vector $\pgt(y)$ is created based on the ground truth label $y$, as defined in Equation \ref{eq:onehot}. 
The final target distribution $p_{\our}(y)$ is then computed using Equation \ref{eq:pour}, which combines the sparsified teacher logits $\tea(x)$ and the one-hot ground truth vector $\pgt(y)$. 
This combination balances the teacher's knowledge with the emphasis on the correct target label. 
The resulting logits-based dataset $(X, Y, P_{\our})$ is then used to fine-tune a student model, leveraging the target distributions for improved knowledge transfer.


\paragraph{Dataset Details.}
\begin{table*}[t]
\centering
\setlength{\tabcolsep}{1.5mm}
\begin{tabular}{lccccccc}
\toprule             
Dataset& GSM8K & MATH  & College &  GaoKao & Minerva  & Olympiad & Average \\
& & & Math &  2023 en & Math  & Bench &  \\
\midrule
Baseline & 46.9 & 31.6 & 18.6 & 26.2 & 5.5 & 7.0 & 22.6 \\
Socratic & 35.9 & 20.3 & 8.8 & 17.4 & 3.7 & 3.4 & 14.9 \\
ScienceQA & 39.7 & 21.6 & 11.7 & 15.8 & 4.4 & 5.9 & 16.5 \\
StackExchange & 37.8 & 22.3 & 12.9 & 19.5 & 3.3 & 4.6 & 16.7 \\
Camel-AI & 41.0 & 22.1 & 11.3 & 20.3 & 5.1 & 3.6 & 17.2 \\
MathInstruct & 40.9 & 24.4 & 12.7 & 20.0 & 6.2 & 4.6 & 18.1 \\
GSM8K & 45.7 & 29.4 & 16.9 & 23.6 & 5.9 & 5.8 & 21.2 \\
MetaMath & \textbf{54.8} & 28.8 & 21.4 & 19.7 & 7.0 & 7.3 & 23.2 \\
\rowcolor{teal!15} Metamath-GSM8K & 54.1 & 29.7 & 21.1 & 24.9 & 4.8 & \textbf{8.1} & 23.8 \\
\rowcolor{teal!15} OpenMathInstruct2 & 49.7 & \textbf{32.7} & \textbf{23.4} & \textbf{27.8} & \textbf{7.4} & 6.7 & \textbf{24.6} \\
\bottomrule
\end{tabular}
\caption{ Results of LLaMA3.2-1b-instruct after supervised fine-tuning on various datasets, including Socratic~\citep{tiger}, StackExchange~\citep{tiger}, Camel-AI~\citep{camel}, MathInstruct~\citep{mathinstruct}, GSM8K~\citep{gsm8k}, MetaMath~\citep{metamath}, MetaMath-GSM8K~\citep{metamath}, and OpenMathInstruct2~\citep{openmathinstruct2}.}
\label{tab:dataset-ablation}
\end{table*}

\Cref{tab:dataset-ablation} presents the results of supervised fine-tuning of LLaMA3.2-1B-Instruct on a variety of mathematical reasoning datasets, including Socratic~\citep{tiger}, StackExchange~\citep{tiger}, Camel-AI~\citep{camel}, MathInstruct~\citep{mathinstruct}, GSM8K~\citep{gsm8k}, MetaMath~\citep{metamath}, MetaMath-GSM8K~\citep{metamath}, and OpenMathInstruct2~\citep{openmathinstruct2}.
Among them, OpenMathInstruct2 demonstrates the strongest overall performance, achieving the highest average score (24.6\%) and outperforming other datasets on most dataset yields competitive performance (23.8\%) and the best result on the Olympiad Bench. These results suggest that datasets like MetaMath-GSM8K and OpenMathInstruct2 can lead to more robust and generalizable mathematical reasoning capabilities. Therefore, our final 1.2M logits dataset consists of 1M samples from MetaMath-GSM8K and 240K from OpenMathInstruct2. More details are shown in \cref{sec:data_ablation_app}. The teacher model utilized for logits generation is LLaMA3.1-70B-Instruct~\citep{llama3}.

\subsection{Finetuning Method}
\label{sec:lft_training}
Using the proposed distribution $p_L$ mentioned above, we fine-tune the student model.

\paragraph{Loss Function.}
Our Logits-based Finetuning (LFT) method uses the Kullback-Leibler (KL) divergence as the loss function to train the student model. The loss function is defined as:
\begin{equation} \label{eq:lour}
L_{\our}(\theta) := \E_{(x, y) \sim (X, Y)} \Big[ \KL\big(p_{\our} | \stu^\theta\big)(y|x)\Big],
\end{equation}
where $x$ and $y$ represent the input and output sequences. $(X, Y)$ is the dataset of input-output pairs. $\E[\cdot]$ denotes the expectation over the dataset.

\paragraph{Fine-tuning.}
This Logits-Based Fine-tune leverages a pre-generated logits dataset, as described in \cref{sec:lft_dataset}, to guide the training of a student model $\stu^\theta$. \cref{alg:lft_training} details our logits-based fine-tuning procedure. For each batch $B$ from the dataset, the student's parameters $\theta$ are updated by minimizing the loss $L_{\our}$ (\cref{eq:lour}), which measures the divergence $\D$ between $p_{\our}$ and $\stu^\theta(y|x)$. This process  results in a trained student model that incorporates knowledge from the teacher logits and ground truth labels.

\begin{figure}[t]
  \includegraphics[width=\linewidth]{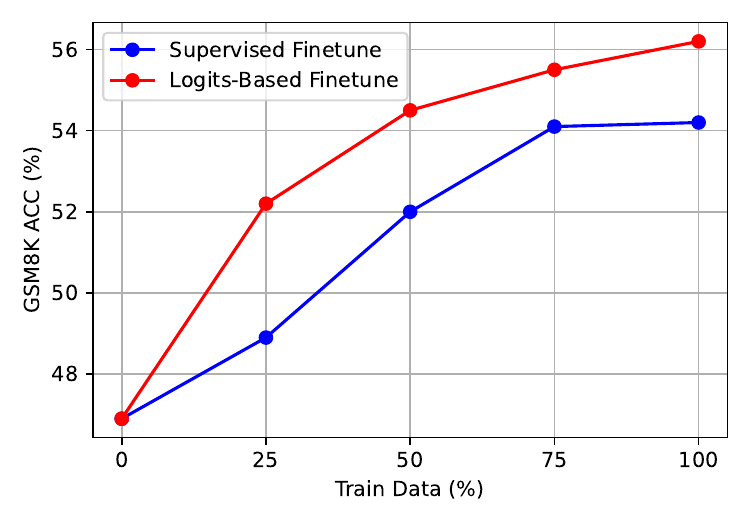}
  \caption{Ablation of our logits-based finetune comparing with baseline trained on different percentage of MetaMath-GSM8K~\cite{metamath} and evaluated on GSM8K~\cite{gsm8k}.}
  \label{fig:method_ablation}
\end{figure}

\begin{table*}[t]
  \centering
  \begin{tabular}{lll}
    \toprule
    \textbf{Data Source} & \textbf{Data Description} & \textbf{Data Count} \\
    \midrule
      & Problem Synthesize from Math & 832k \\
    OpenMathInstruct & Problem Synthesize from GSM8K & 138k \\
    \cite{openmathinstruct2} & CoT Aug from Math & 15k \\
      & CoT Aug from GSM8K & 15k \\
    \midrule
      & Answer Aug from GSM8K & 80k \\
     MetaMath & Rephrasing from GSM8K & 80k \\
     \cite{metamath} & Self-Verification from GSM8K & 40k \\
      & Forward-Backward Reasoning from GSM8K & 40k \\
    \bottomrule
  \end{tabular}
\caption{ Source and description of our 1.2M logits dataset, including 240K from MetaMath-GSM8K~\cite{metamath}, and 1M from OpenMathInstruct2~\cite{openmathinstruct2}.}
\label{tab:logits_dataset_description}
\end{table*}
\begin{table}
  \centering
\begin{tabular}{lc}
\toprule
\textbf{} & \textbf{GSM8K} \\ 
\midrule
LLaMA3.2-1b-It & 46.9 \\ 
Supervsied Finetune & 54.1 \\ 
\rowcolor{teal!15} Logits-based Finetune & \textbf{56.1} \\  
\bottomrule
\end{tabular}
\caption{ 
    Ablation of our logits-based finetune comparing with baseline trained on MetaMath-GSM8K~\cite{metamath}.
  }
  \label{sec:ablation}
\end{table}

\begin{table*}[t]
  \centering
  \setlength{\tabcolsep}{0.7mm}
  \begin{tabular}{lcccccccccccc}
    \toprule
Model & GSM8K & MATH  & College & Olympiad & Svamp & ASDiv & Mawps & Carp & TabMWP & Avg \\
& & & Math  & Bench & & & & en &  & \\
\midrule
Gemma-2-2b-It & 61.9 & 26.1 & 20.6 & 5.3 & 68.7 & 77.6 & 89.7 & 32.7 & 42.7 & 47.26 \\
Phi-3.5-Mini-It & 87.2 & 45.2 & 35.9 & 12.3 & 83.7 & 85.9 & 88.1 & 35.1 & 55.7 & 58.79 \\
LLaMA3.2-1b-It & 46.9 & 31.6 & 18.6 & 7.0 & 69.3 & 70.0 & 79.3 & 30.5 & 33.4 & 42.96 \\
LLaMA3.2-3b-It& 81.3 & 51.7 & 34.1 & 17.2 & 86.4 & 89.0 & 96.7 & 45.1 & 70.0 & 63.50 \\
\midrule
\rowcolor{teal!15} \ourmodel-1b & 55.0 & 35.1 & 25.3 & 7.6 & 72.5 & 78.5 & 87.3 & 34.8 & 56.1 & 50.24 \\
\rowcolor{teal!15} \ourmodel-3b & 81.0 & 51.3 & 36.3 & 16.1 & 88.4 & 90.6 & 96.3 & 46.0 & 74.3 & 64.48 \\
    \bottomrule
  \end{tabular}
\caption{
Performance of our ScienceLLaMA comparing with current SOTAs on various benchmarks, including Socratic~\citep{tiger}, StackExchange~\citep{tiger}, Camel-AI~\citep{camel}, MathInstruct~\citep{mathinstruct}, GSM8K~\citep{gsm8k}, MetaMath~\citep{metamath}, MetaMath-GSM8K~\citep{metamath}, and OpenMathInstruct2~\citep{openmathinstruct2}.}
\label{tab:performance}

\end{table*}

\section{Experiment}
In this section, we present a comprehensive evaluation of our logits-based fine-tuning approach. We first detail our evaluation benchmarks  in~\cref{sec:benchmark} and training details in~\cref{sec:implement_details}. Then, we analyze key components through ablation studies in~\cref{sec:ablation}. Finally, we compare on multiple datasets against baselines in~\cref{sec:performance}.

\subsection{Benchmark} 
\label{sec:benchmark}
We evaluate our \ourmodel on mathematical benchmark including: 

\para{GSM8K} (Grade School Math 8K,~\citet{gsm8k}) is a dataset comprising 8.5K high-quality, linguistically diverse grade school math word problems.

\para{MATH}~\citep{math} consists of 12,500 challenging competition mathematics problems, each accompanied by a detailed step-by-step solution.

\para{OlympiadBench}~\citep{olympiadbench} presents an Olympiad-level bilingual multimodal scientific benchmark with 8,476 problems from challenging mathematics and physics competitions like the Chinese college entrance exam.

\para{CollegeMath}~\citep{collegemath} is a mathematical reasoning dataset created using MathScale, containing two million math question-answer pairs.

\para{SVAMP} (Simple Variations on Arithmetic Math word Problems,~\citet{svamp}) introduces a challenge dataset for English math word problems.

\para{ASDiv}  (Academia Sinica Diverse MWP Dataset,~\citet{asdiv}) offers a diverse English math word problem corpus consisting of 2,305 problems,.

\para{MAWPS}  (MAth Word ProblemS,~\citet{mawps}) is an online repository providing a unified testbed to evaluate algorithms on Math Word Problems.

\para{CarpEN} (Computation-intensive AlgebRa Problems,~\citet{carp}) constructs a Chinese dataset focused on computation-intensive algebra problems.

\para{TabMWP} (Tabular Math Word Problems,~\citet{tabmwp}) contains 38,431 open-domain grade-level math problems requiring reasoning over textual and tabular data.

\subsection{Training Details}
\label{sec:implement_details}
We train the LLaMA3.2-1B/3B-Instruct as our model  on our constructed 1.2M science logits dataset using our proposed logits-based fine-tuning method. The resulting trained models are referred to as ScienceLLaMA-1B/3B. We set the batch size to 1 and the learning rate to $2\times 10^{-5}$. All experiments are conducted on 8 Nvidia A800 GPUs.

\subsection{Ablation}
\label{sec:ablation}
\Cref{fig:method_ablation} presents the GSM8K accuracy  of our logits-based fine-tuning in comparison to supervised fine-tuning, trained on varying percentages of the MetaMath-GSM8K dataset and evaluated on the GSM8K benchmark. Both methods demonstrate improved performance as the proportion of training data increases, but the logits-based fine-tuning consistently outperforms supervised fine-tuning across all data scales. Notably, the accuracy achieved by the logits-based approach with just 25\% of the training data exceeds that of the supervised method trained on 50\% of the data. Furthermore, with half of the training data, the logits-based approach achieves better results than the supervised method trained on the full dataset. On the complete training set, our logits-based fine-tuning achieves an accuracy of 56.1\%, surpassing the supervised fine-tuning baseline by 2.0\% and outperforming the original pre-trained model by 9.2\%. These findings underscore the effectiveness of leveraging logits to guide the learning process.

\subsection{Performance}
\label{sec:performance}
As shown in~\cref{tab:performance}, we evaluate our proposed method on various math benchmarks. Our ScienceLLaMA significantly outperforms the SFT model. Specifically, the ScienceLLaMA-1B model surpasses the directly SFT-trained LLaMA3.2-1B-Instruct on Mawps and TabMWP by 18\% and 22.7\% in accuracy, respectively. Furthermore, for the average score across nine benchmarks, our ScienceLLaMA-1B achieves a 7.28\% higher accuracy. These results demonstrate that our method exhibits strong stability and generalization, significantly outperforming the Supervised-Finetuning approach. 

\section{Related Works}
\paragraph{Large Language Models.} 
Recently, LLMs have demonstrated remarkable capabilities across a wide range of tasks~\cite{gpt3,lamda,palm,gpt4,palm2, yang2024visionzip}, including machine translation~\cite{robocoder}, text summarization~\cite{cape}, dialogue generation~\cite{instructgpt}, and code generation~\cite{motcoder}. While their capacity is impressive, these advanced abilities often emerge only in models with substantial parameter sizes~\citep{scaling_law,emergent}, which demand significant computational resources. As a result, model compression has become essential to facilitate the practical deployment of LLMs and to support further research in the field.

\paragraph{Knowledge Distillation.}
Knowledge Distillation (KD;~\citet{kd}), a popular model compression method, focuses on training a smaller student model under the guidance of a larger teacher model~\citep{policy_kd,distilbert,kdsurvey}. In NLP, KD has been widely applied to classification tasks by replicating the teacher model's output distribution~\citep{lightpaff,mixkd,perturb_kd}, internal layer representations~\citep{tinybert,bert-pkd}, or attention patterns~\citep{minilm,minilmv2}. For text generation tasks, traditional KD typically minimizes the Kullback-Leibler divergence (KLD) between the teacher's and student's output distributions, using the teacher’s output as supervision at every time step~\citep{distilbert} or directly training the student on text sequences generated by the teacher~\citep{skd,alpaca,vicuna,ITGPT4}. Unlike recent studies~\citep{gkd,f-div-kd,distillm,minillm}, which focus on alternative distribution discrepancy metrics in KD, our work emphasizes the creation of a distribution that integrates the robustness of the ground truth with the teacher's token-level knowledge priors.

\section{Conclusion}
In this work, we address the limitations of traditional supervised fine-tuning for developing compact and efficient LLMs by introducing a novel logits-based fine-tuning framework. Our approach integrates the strengths of supervised learning and knowledge distillation, constructing enriched training targets that combine teacher logits with ground truth labels. This method preserves both correctness and linguistic diversity, enabling the student model to learn from the teacher’s knowledge while maintaining task-specific supervision. We constructed a large-scale 1.2M science logits dataset and trained a series of science-focused models, referred to as ScienceLLaMA. Experimental results demonstrate that our method achieves significant improvements over state-of-the-art supervised fine-tuning approaches, with accuracy gains of 18\% on Mawps and 22.7\% on TabMWP. Across nine widely used mathematical benchmarks, our method consistently outperforms prior SFT models, achieving an average improvement of 7.28\%. These results highlight the robustness of our logits-based fine-tuning framework.

\clearpage
\section*{Limitations}
While our work successfully introduces a distillation framework tailored for large language models (LLMs) using a logits-based instruction tuning strategy, our experiments were constrained by computational resources, limiting the scale of the evaluated models. We plan to extend this approach to larger model architectures in future work.

\section*{Broader Impact}
By refining the distillation process to better preserve the teacher model’s reasoning capabilities, our method may enable more compact and deployable models. This could make LLM-powered applications—such as real-time conversational assistants, on-device AI tools, and resource-constrained edge computing—more accessible and practical. 
 However, the broader deployment of efficient, distilled models also introduces risks. If misused, malicious actors might exploit distillation techniques to create highly optimized models for harmful purposes, such as generating convincing misinformation or automating fraudulent interactions. Responsible development and rigorous evaluation frameworks will be essential to mitigate these risks while maximizing the societal benefits of our method.

\section*{AI Assistance Disclosure}
In the preparation of this work, the authors used large language models (LLMs) for writing assistance during manuscript composition. Following initial drafting, the authors reviewed and edited the content as needed and take full responsibility for the final publication.

\bibliography{custom}

\appendix

\label{sec:appendix}

\begin{table}[t]
\centering
\small
\label{tab:token_probs}
\begin{tabular}{lllll}
\cellcolor{teal!25}There & \cellcolor{teal!25}\textbackslash n\textbackslash n &  &  & \\
\cellcolor{teal!47} are & \cellcolor{teal!2} is & \cellcolor{teal!0}'s &  & \\
\cellcolor{teal!49}<space> & \cellcolor{teal!1} two &  &  & \\
\cellcolor{teal!45}12 & \cellcolor{teal!2}6 & \cellcolor{teal!2}3 & \cellcolor{teal!1}2 & \\
\cellcolor{teal!50} inches &  &  &  & \\
\cellcolor{teal!48} in & \cellcolor{teal!1} per &  &  & \\
\cellcolor{teal!34}<space> & \cellcolor{teal!8} a & \cellcolor{teal!4} each & \cellcolor{teal!3} one & \cellcolor{teal!1} every\\
\cellcolor{teal!50}1 &  &  &  & \\
\cellcolor{teal!50} foot &  &  &  & \\
\cellcolor{teal!29}, & \cellcolor{teal!15}. & \cellcolor{teal!4}.\textbackslash n & \cellcolor{teal!1}.\textbackslash n\textbackslash n & \cellcolor{teal!0} so\\
\cellcolor{teal!48} so & \cellcolor{teal!1} and & \cellcolor{teal!1} therefore & \cellcolor{teal!0}<space> & \\
\cellcolor{teal!40} Vlad & \cellcolor{teal!4} we & \cellcolor{teal!2}<space> & \cellcolor{teal!2} multiply & \cellcolor{teal!2} convert\\
\cellcolor{teal!29}'s & \cellcolor{teal!21} is & \cellcolor{teal!0}’s &  & \\
\cellcolor{teal!49} height & \cellcolor{teal!1}<space> & \cellcolor{teal!0} total &  & \\
\cellcolor{teal!30} is & \cellcolor{teal!13} in & \cellcolor{teal!5} can & \cellcolor{teal!1} of & \cellcolor{teal!1},\\
\cellcolor{teal!38}<space> & \cellcolor{teal!6} \textbackslash \textbackslash ( & \cellcolor{teal!3}\textbackslash n & \cellcolor{teal!2} \$ & \cellcolor{teal!2} (\\
\cellcolor{teal!48}6 & \cellcolor{teal!1}12 &  &  & \\
\cellcolor{teal!29} * & \cellcolor{teal!10} feet & \cellcolor{teal!5}( & \cellcolor{teal!5} x & \cellcolor{teal!2} \textbackslash \textbackslash \\
\cellcolor{teal!50}<space> & \cellcolor{teal!0}12 &  &  & \\
\cellcolor{teal!50}12 &  &  &  & \\
\cellcolor{teal!45} + & \cellcolor{teal!5} = &  &  & \\
\cellcolor{teal!50}<space> &  &  &  & \\
\cellcolor{teal!50}3 &  &  &  & \\
\cellcolor{teal!47} = & \cellcolor{teal!2} inches &  &  & \\
\cellcolor{teal!50}<space> &  &  &  & \\
\cellcolor{teal!44}75 & \cellcolor{teal!6}72 & \cellcolor{teal!0}63 &  & \\
\cellcolor{teal!49} inches & \cellcolor{teal!0} in &  &  & \\
\cellcolor{teal!26}.\textbackslash n & \cellcolor{teal!13}. & \cellcolor{teal!5} and & \cellcolor{teal!5}, & \cellcolor{teal!0}.\textbackslash n\textbackslash n\\
\cellcolor{teal!41}His & \cellcolor{teal!3}V & \cellcolor{teal!2}Similarly & \cellcolor{teal!2}The & \cellcolor{teal!2}S\\
\cellcolor{teal!48} sister & \cellcolor{teal!1} younger & \cellcolor{teal!0} sisters &  & \\
\cellcolor{teal!43}'s & \cellcolor{teal!6} is &  &  & \\
\cellcolor{teal!50} height &  &  &  & \\
\cellcolor{teal!50} is &  &  &  & \\
\cellcolor{teal!50}<space> &  &  &  & \\
\cellcolor{teal!50}2 &  &  &  & \\
\cellcolor{teal!49} * & \cellcolor{teal!1}* &  &  & \\
\cellcolor{teal!50}<space> &  &  &  & \\
\cellcolor{teal!50}12 &  &  &  & \\
\cellcolor{teal!50} + &  &  &  & \\
\cellcolor{teal!50}<space> &  &  &  & \\
\cellcolor{teal!50}10 &  &  &  & \\
\cellcolor{teal!50} = &  &  &  & \\
\cellcolor{teal!50}<space> &  &  &  & \\
\cellcolor{teal!50}34 &  &  &  & \\
\cellcolor{teal!50} inches &  &  &  & \\
\cellcolor{teal!44}.\textbackslash n & \cellcolor{teal!3}. & \cellcolor{teal!2}, & \cellcolor{teal!0}.\textbackslash n\textbackslash n & \\
\end{tabular}
\caption{Example of the logits-based label of \textit{There are 12 inches in 1 foot, so Vlad's height is 6 * 12 + 3 = 75 inches.
His sister's height is 2 * 12 + 10 = 34 inches.}.}
\label{tab:logits-label-example}
\end{table}

\begin{table*}[t]
\centering
\setlength{\tabcolsep}{1mm}
\begin{tabular}{llll}
\toprule
\textbf{Dataset} & \textbf{Sampled Size} & \textbf{Data Source} & \textbf{Paper Source} \\
\midrule
Socratic & 511k & \href{https://huggingface.co/datasets/TIGER-Lab/WebInstructSub}{\texttt{TIGER-Lab/WebInstructSub}} & \citet{tiger} \\
StackExchange & 291k & \href{https://huggingface.co/datasets/TIGER-Lab/WebInstructSub}{\texttt{TIGER-Lab/WebInstructSub}} & \citet{tiger} \\
ScienceQA & 100k & \href{https://huggingface.co/datasets/ibivibiv/science_qa}{\texttt{ibivibiv/science\_qa}} & - \\
\multirow{4}{*}{Camel-AI} & 50k & \href{https://huggingface.co/datasets/camel-ai/math}{\texttt{camel-ai/math}} & \citet{camel} \\
& 20k & \href{https://huggingface.co/datasets/camel-ai/physics}{\texttt{camel-ai/physics}} & \citet{camel} \\
& 20k & \href{https://huggingface.co/datasets/camel-ai/biology}{\texttt{camel-ai/biology}} & \citet{camel} \\
& 20k & \href{https://huggingface.co/datasets/camel-ai/chemistry}{\texttt{camel-ai/chemistry}} & \citet{camel} \\
MathInstruct & 262k & \href{https://huggingface.co/datasets/TIGER-Lab/MathInstruct}{\texttt{TIGER-Lab/MathInstruct}} & \citet{mathinstruct} \\
GSM8K & 7.5k & \href{https://huggingface.co/datasets/openai/gsm8k}{\texttt{openai/gsm8k}} & \citet{gsm8k} \\
MetaMath & 395k & \href{https://huggingface.co/datasets/meta-math/MetaMathQA}{\texttt{meta-math/MetaMathQA}} & \citet{metamath} \\
MetaMath-GSM8K & 240k & \href{https://huggingface.co/datasets/meta-math/MetaMathQA}{\texttt{meta-math/MetaMathQA}} & \citet{metamath} \\
OpenMathInstruct2 & 10k & \href{https://huggingface.co/datasets/nvidia/OpenMathInstruct-2}{\texttt{nvidia/OpenMathInstruct-2}} & \citet{openmathinstruct2} \\
\bottomrule
\end{tabular}
\caption{Size and Source of the datasets, including Socratic~\citep{tiger}, StackExchange~\citep{tiger}, Camel-AI~\citep{camel}, MathInstruct~\citep{mathinstruct}, GSM8K~\citep{gsm8k}, MetaMath~\citep{metamath}, MetaMath-GSM8K~\citep{metamath}, and OpenMathInstruct2~\citep{openmathinstruct2}.}
\label{tab:detailed_datasets_ablation}
\end{table*}

\section{Verification of Constraints}
\label{sec:demonstration}
We now demonstrate that the proposed distribution $p_{\our}(y_i)$ satisfies the four constraints in \cref{sec:constraints}.

\paragraph{Constraint 1.} Since $\tea(y_i)\in [0,1]^M$, and $\pgt(y_i)$ is a one-hot vector with a value of 1 at index $y_i$ and 0 elsewhere, the largest value in $\tea(y_i) + \pgt(y_i)$ will always be at index $y_i$. The normalization by the L1 norm preserves this relationship, ensuring $p_{\our {y_i}}(y_i) \ge p_{\our j}(y_i)$ for all $j \ne y_i$. Thus, Constraint 1 is satisfied.

\paragraph{Constraint 2.}  This constraint pertains to the relative proportions within the Top $K$ elements of $\tea(y_i)$. Since $\pgt(y_i)$ only modifies the ground truth index, and the normalization factor is applied uniformly across all elements, the relative proportions among the other Top-K elements remain unchanged. Specifically, for $j, k \in \text{Top-K}(y_i)$ and $j, k \neq y_i$, we have:
\begin{equation}
\begin{aligned}
    \frac{p_{\our_j}(y_i)}{p_{\our_k}(y_i)} &= \frac{(\tea(y_i)_j + 0)/\|\tea(y_i)+\pgt(y_i)\|_1}{(\tea(y_i)_k + 0)/\|\tea(y_i)+\pgt(y_i)\|_1} \\&= \frac{\tea(y_i)_j}{\tea(y_i)_k}.
\end{aligned}
\end{equation}
If $y_i$ is within the Top $K$, the ratio involving $y_i$ also holds due to the uniform scaling by the L1 norm. Therefore, Constraint 2 is satisfied.

\paragraph{Constraint 3.} For any $j \notin S$, $\tea(y_i)_j = 0$ (due to Top $K$ sparsification). Therefore, $p_{\our_j}(y_i) = 0$. For any $k \in S$, $p_{\our_k}(y_i)$ will be non-negative due to either a non-zero value in $\tea(y_i)$ or the one-hot vector $\pgt(y_i)$. Therefore, $p_{\our_j}(y_i) \le p_{\our_k}(y_i)$ for all $j \notin S$ and $k \in S$, satisfying Constraint 3.

\paragraph{Constraint 4.} By definition, the L1 norm normalization in Equation \ref{eq:pour} ensures that the elements of $p_{\our}(y_i)$ sum to 1. Furthermore, since both $\tea(y_i)$ and $\pgt(y_i)$ have non-negative elements, $p_{\our}(y_i)$ will also have non-negative elements. The normalization then ensures that all elements are within the range [0, 1]. Thus, Constraint 4 is satisfied.

\section{Dataset Details}
\label{sec:data_ablation_app}
\Cref{tab:detailed_datasets_ablation} provides a comprehensive overview of the datasets used in our study, detailing their sampled sizes, data sources, and associated references. The datasets include Socratic and StackExchange from \cite{tiger}, Camel-AI (covering math, physics, biology, and chemistry) from \cite{camel}, MathInstruct from \cite{mathinstruct}, GSM8K from \cite{gsm8k}, MetaMath and MetaMath-GSM8K from \cite{metamath}, and OpenMathInstruct2 from \cite{openmathinstruct2}. For OpenMathInstruct2, which contains 1M samples, we sampled 10K for evaluation.

\section{Logits-Based Dataset Example}
\Cref{tab:logits-label-example} presents a logits-based label visualization for the sentence:
"There are 12 inches in 1 foot, so Vlad’s height is $6 \times 12 + 3 = 75$ inches. His sister’s height is $2 \times 12 + 10 = 34$ inches."

\end{document}